\documentclass[letterpaper, 10 pt, conference]{ieeeconf}  
                                                          
\IEEEoverridecommandlockouts                              
\usepackage{comment}
\usepackage{graphicx}
\graphicspath{ {./images/} }
\usepackage{subcaption}

\title{\LARGE \bf
Tuned Inception V3 for Recognizing States of Cooking Ingredients
}

\author{Kin NG\\
Computer Science and Engineering Department\\
University of South Florida
}

\begin{document}

\maketitle
\thispagestyle{empty}
\pagestyle{empty}

\begin{abstract}

Cooking is a task that must be performed in a daily basis, and thus it is an activity that many people take for granted. For humans preparing a meal comes naturally, but for robots even preparing a simple sandwich results in an extremely difficult task. In robotics, designing kitchen robots is complicated since cooking relies on a variety of physical interactions that are dependent on different conditions such as changes in the environment, proper execution of sequential instructions, along with motions, and detection of the different states in which cooking-ingredients can be in for their correct grasping and manipulation. In this paper, we focus on the challenge of state recognition and propose a fine tuned convolutional neural network that makes use of transfer learning by reusing the Inception V3 pre-trained model. The model is trained and validated on a cooking dataset consisting of eleven states (e.g. peeled, diced, whole, etc.). The work presented on this paper could provide insight into finding a potential solution to the problem.

\end{abstract}

\section{INTRODUCTION}

In the past few years, deep learning has achieved great success, as well as, contributed improvements in computer vision related challenges. One clear example is object recognition. With the introduction of very deep convolutional neural networks (CNN) \cite{c11}, object classification and recognition tasks do not pose a significant challenge anymore in the area of computer vision \cite{c10}. The successful trend of using CNN for image analysis started with AlexNet \cite{c10}, a deep CNN capable of acquiring a 15.3\% error rate in the ImageNet large scale visual recognition challenge. Since then, different CNN architectures have been proposed such as, VGGNet \cite{c11}, Google Net \cite{c21}, Inception V3 \cite{c12}, RestNet \cite{c13}, among others. Each of these models have been able to constantly reduce the error rate for object detection significantly, to the point that nowadays a machine is capable to obtain a precision that is extremely close, and arguably better, than human performance. 

The significant improvements in computer vision-related challenges has inspired scientists to use deep learning as the starting point to solve common problems in robotic applications. For example, using deep reinforcement learning to teach machines to be able to interact with their surroundings, recognize strong changes in the environment, and detect the location of objects, has shown to be promising in the field of autonomous driving, and it is representing a step forward in the right direction for automotive applications \cite{c22, c23}.

In addition, throughout the years, the field of robotics has seen great success in a variety of areas. In factories, they are used to carry out tasks that can be described as dangerous or too complex for humans. In the automobile industry, to weld parts on auto assembly lines, which speeds up the process of making cars by introducing automation. In the medical field \cite{c24}, robot applications play a key role in surgical interventions, as well as, assistance in rehabilitation. 

However, a common limitation these robotic applications face is that they are programmed to function in a certain way, which means that these tasks only involve following a repeated series of sequential actions that are not dependent on variable factors such as changes in the environment, or the various conditions and different states an object can be in at certain periods of time. On the contrary, these are essential factors for accomplishing daily-living activities such as housework, transportation, or meal preparation. Thus, the challenge of designing robots with the goal of helping us with the aforementioned tasks is starting to receive a significant amount of attention. To achieve this, robots must be capable of understanding and recognizing their surroundings, as well as, having a fundamental knowledge of proper object grasping and manipulation based on the current state the object is in. For instance, cooking is a daily-live activity that could give us some insight on how to tackle these challenges, and consequently it is the main motivation for this paper.

In this paper, we will focus on the state recognition challenge, which consists of identifying the state in which a cooking ingredient is currently in. For example, in order to make french fries, our ingredient, in this case a potato, will go through a series of transformations such as, whole, peeled, chopped, and julienne, before we are able to fry it. This highlights the importance of being able to recognize the different states of our ingredients since their grasping and manipulation are dependent on these.

This paper is organized as follows: section 2 presents previous work related to this challenge. Section 3 gives a brief description of the dataset used for this challenge, including how it was collected, annotated, and preprocessed. Section 4 discusses the methodology and introduces the architecture proposed in this paper. Section 5 shows the evaluation of our proposed architecture, including the different training parameters and methods used, and we present the results of our proposed model on unseen test data. Finally, in section 6, we discuss our findings and conclude the paper.

\section{Related Work}

Eating is a basic necessity for living since it provides our body with the energy it needs to get us through the day. Thus, cooking becomes a task that most of us must perform every single day. Similarly, it is also one of the tasks that consumes most of our time as on average it needs to be performed at least three times per day. Hence, by designing a robot cooking assistant capable of preparing meals for us, we would be able to spend our time more efficiently on other productive activities. Consequently, it would be helpful for elderly and disable people, who are not able to cook by themselves, as well as, people not skilled at cooking, who struggle to prepare even the simplest dishes. 

However, as natural as it may result for humans, cooking still presents a great challenge for robots because of the numerous factors it depends on. For a robot to be capable of successfully preparing a dish, it must know the steps that it takes for preparing it, the grasping and manipulation methods that are needed while cooking, and the different transformations and states that each ingredient can undergo.

In recent years, there have been several research on cooking-related tasks that could provide valuable insight for the cooking challenge. In \cite{c19} and \cite{c7}, two interesting approaches for cooking-recipe retrieval based on ingredient recognition in food-dishes are proposed. Both experiments use deep-based CNN architectures to tackle the challenge. In \cite{c7}, the zero-shot retrieval problem is presented, which consists in the inability of retrieving cooking recipes with an unknown food category. On the other hand, \cite{c19} tackles the problem of cross-modal retrieval, which is retrieving images of a dish by providing a recipe, or vice-versa. 

Another interesting work based on functional motion was presented in \cite{c14, c15, c16}, in which a novel functional object-oriented network (FOON) was proposed. This structured knowledge representation was learned based on the observations of human activities and manipulations with different objects and its changes of state. The main purpose of FOON is to create a sequence of executable steps to solve manipulation and grasping related-problems, which were focused around cooking activities and recipes, but can also be generalized to a broader range of tasks. 

Furthermore, Sun et al. presented an interesting dataset of daily interactive manipulations focused on the position, orientation, torque and force of manipulated objects \cite{c3}. The proposed dataset could give us valuable insight on the different parameters needed to manipulate certain objects, and could help to design robots capable of adapting their way of grasping objects based on the motions to be executed. Some examples of cooking-related motions included in the dataset are stirring, pouring, mixing, and slicing. 

In \cite{c9}, a dataset of objects, scenes and materials was introduced to study the different physical world transformations that these can undergo. The paper focused on understanding object states and discovering the transformations to and from those states by using a novelty approach, in which they create a chain of images that represent the smooth transitions between one state to the other. 

Recently, some work has been conducted regarding the object state detection challenge. In \cite{c6} and \cite{c2}, two different approaches to solve this problem were presented. \cite{c6} used a fine tuned VGG16 pre-trained model, while \cite{c2} proposed a modified Inception V3 model. Both approaches achieved promising accuracy of 77\% and 73\%, respectively, for unseen test data. However, in these two papers, models were trained with version 1.0 of the cooking dataset found in \cite{c1}, which was limited to only 7 states. Recently, 4 more states have been added to this dataset. Thus, a new deep-based CNN model needs to be designed and trained to properly account for these new states. In this paper, we focused on fine-tuning and modifying the pre-trained Inception V3 model to study its impact and performance on the object state recognition challenge.

\section{Data Collection and Preprocessing}

The dataset for this study was crawled from Google, and consists of cooking images that were manually labeled according to its state. At this moment, version 1.2 of the dataset consists of 9309 images, and we can find a total of 11 classes related to these images (creamy/paste, diced, floured, grated, juiced, julienne, mixed, other, peeled, sliced, and whole). In \cite{c1}, we can find a in depth analysis of the dataset, as well as, a detailed description of what each state represents. In figure 1, we present some examples of the images on our dataset and their respective state.

\begin{figure*}
\centering
\includegraphics[width=\textwidth]{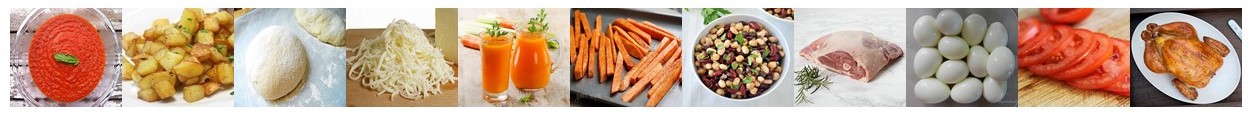}
\caption{Cooking states, from left to right: creamy/paste, diced, floured, grated, juiced, julienne, mixed, other, peeled, sliced, and whole}
\end{figure*}

\subsection{Data Annotation}

In addition to crawled images from Google, a part of the dataset also consists of images that were collected through data annotation. Some students from the University of South Florida, were given the task of annotating batches of approximately 5 to 20 minute long cooking-related videos. Students were assigned the task of keeping track of a specific cooking ingredient appearing in the video and labeling it according to the state it was in. Annotations were checked and verified to make sure that the labeling was correct. Furthermore, all state annotations were performed using Vatic annotation tool \cite{c27}.

\subsection{Data Pre-processing}

The dataset was split in the following way: 68\% for training (6,348 images), 15\% for validation (1377 images), and 17\% testing (1584 images). The input shape of our images was set to (299x299) since this is the default shape used by Inception V3 model. In addition, We applied the following pre-processing techniques to our dataset to allow for smoother and faster training. Image re-scaling to fall in the range of 0-1, sample-wise centering and normalization, which normalizes the mean value of each data sample. Lastly, ZCA whitening is also performed to highlight important features and structures of an image, as well as, reducing the redundancy of some of the pixels in the image.

\subsection{Data Augmentation}

For the state recognition challenge, one of the limitations we encounter with our current dataset is its size. If we desire to achieve a good performance with our model, then it is important to have a large enough dataset that we can feed to it. Thus, data augmentation plays a key role on the optimization of our model by preventing our network to learn irrelevant patterns by augmenting our data, which ultimately boosts the overall performance of our model and helps to prevent overfitting.

Augmentation allow us to generate more data for our model by performing alterations to our current images. In addition, it exposes our model to learn a variety of conditions in which an object can exist in, such as, different scales, orientations, or illuminations. This results in a more robust and generalized model. In our experiments, images were augmented by applying the following transformations, horizontal and vertical flips, 20\% shearing, 20\% width and height shifting, 20\% zooming, and 30-degree rotation.  

\section{Methodology}

The first stage of our experimentation consisted in testing which pre-trained model will be most suitable for the state recognition challenge. Therefore, we investigated two different pre-trained CNN architectures, VGG19 and Inception V3. In order to test their performance on our collected dataset, we only took into consideration their base models by removing the top classification layers, and replacing them with our own classifier for 11 cooking-related states. Based on both loss and accuracy metrics during testing, we found that Inception V3 performed slightly better than VGG19 network. Therefore, we decided to carry out our experiments further with the network that gave us the best performance. 

\subsection{Inception V3 Proposed Model}
Our proposed model, shown in figure 2, is composed of a convolutional layer, with a size of 1024 neurons. Since the base model for Inception V3 is already very deep because of its numerous blocks, we decided to use a (3x3) convolution mask with the purpose of avoiding a higher convolutional cost just as it was indicated in \cite{c2}. In the next layer, we applied batch normalization to achieve hyper-parameter robustness and avoid the vanishing gradient problem by having an optimal initialization based on Gaussian distribution. In \cite{c18}, it is shown that bath normalization allows for well-behaved and predictive gradients in the training process, which ultimately enables a more effective optimization. An activation layer was also added with a size of 1024 neurons. In addition, we decided to also add a Global Average Pooling (GAP) layer to reduce the size dimensions of the previous layer. According to \cite{c17}, GAP not only is able to reduce dimensions, but it can also act as a regularizer being able to prevent some overfitting for the overall network structure. Before our 11 states classifier, we also added a Dropout layer with a 0.3 dropping rate. In \cite{c5}, it was shown that dropout helps the network to have a lower sensitivity to specific weights in the network, and results in better generalization, which reduces the likelihood for overfitting.

\begin{figure}[ht]
\centering
\includegraphics[width=4cm, height=10cm]{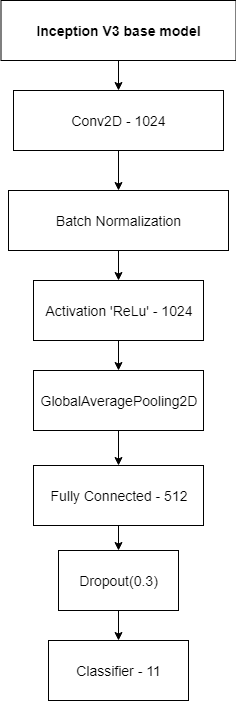}
\caption{Fine-tuned Inception V3 Architecture}
\end{figure}

\section{Evaluation and Results}

In this section, we explore the different training parameters, optimizers, and regularization methods that were used during experimentation, as well as, show the performance of our models in terms of both validation loss and accuracy metrics observed during training. Lastly, we report the results of our final proposed model.

\subsection{Model Training}
Our experiment consisted in a two-stage training process. The first stage trained the top layers of our model, which consists of the layers added on top of the Inception V3 base model, and our 11 states classifier. The purpose of this stage was to properly train the randomly initialized weights on these top layers up to a satisfactory performance and stability before moving on to the second stage. In addition, the base model layers of Inception V3 were frozen since these weights have already been well-trained on Imagenet dataset. This first stage is important as it allows our network to start the fine tuning process with trained weights rather than a mixture of pre-trained and random weights.

Once the top layer weights are well-trained, we move to the second stage of the training process. This stage consists in exploring the effects of fine tuning in our model's performance. In the fine tuning process, we unfreeze some layers from our base model and make their pre-trained weights trainable. From our experiments, we discovered that fine tuning the top two inception blocks (freezing layers up to the 249th layer), improved the overall accuracy of our model. In both stages, we maintain the weights at lower layers of the Inception V3 model frozen since these lower convolutional blocks extract basic features of images such as, shapes, edges, contours, textures, etc. that are also relevant to our object state classification challenge. 

In addition, our experiments made use of early stopping technique to carefully check for overfitting. For both stages, we check the validation loss with a patience of 5 epochs. Training was stopped if no improvement was found within these 5 epochs, and we immediately begin the next stage. Furthermore, the weights that reached the lowest validation loss during training were selected. 

\subsection{Evaluation}

In this portion of the paper, we discuss the different training options that were experimented during the evaluation of our model, compare how different batch sizes, optimizers, learning rates, and regularization methods affected the performance of our model, and present the results that showed a significant improvement. 

\subsubsection{Batch Size}

We experimented with two different batch sizes, 16 and 32. For both experiments, we used Adam optimizer. In table I, we can observe that our model was able to reach a better performance with batch size 32. A larger batch size allows to reduce the variance of gradient updates, and enables a faster progress since we can take bigger-step sizes. For future experiments, we would test our model with bigger batch sizes to check if their effects on the overall performance of our model will lead to positive results.

\begin{table}[h]
\caption{Effect on batch sizes}
\label{table1}
\begin{center}
\begin{tabular}{|c|c|c|}    
\hline
Batch size & Validation Loss & Validation Acc.\\
\hline
16 & 0.8844  & 0.7086\\
\hline
32 & 0.8192  & 0.7286 \\
\hline
\end{tabular}
\end{center}
\end{table}

\subsubsection{Regularization methods}

For our experiments, we decided to test two different regularization methods, dropout and L2 regularizer. Both methods were tested using SGD optimizer. In figure 3, we can observe that using dropout resulted in less overfit for our model than using L2 regularizer, and ultimately enabled our model to be more robust and generalized better. We found experimentally that the dropout rate that works best with our model was 0.3.

\begin{figure}[h]
    \centering
    \begin{subfigure}[h]{0.22\textwidth}
        \includegraphics[height=1.1in]{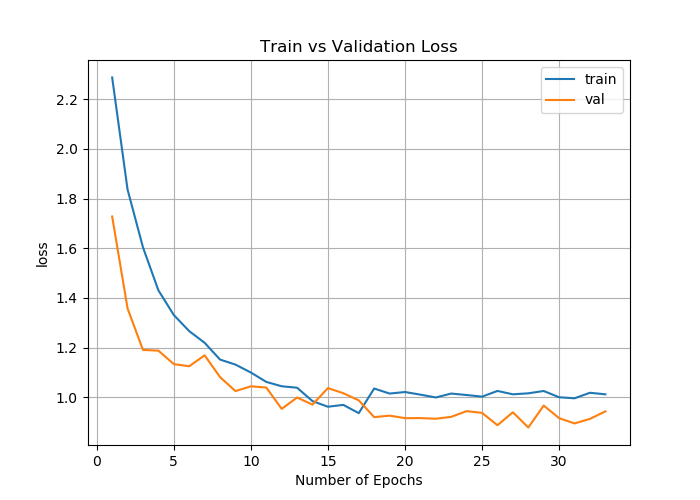}
        \caption{Loss based Dropout}
    \end{subfigure}%
    ~ 
    \begin{subfigure}[h]{0.22\textwidth}
        \includegraphics[height=1.1in]{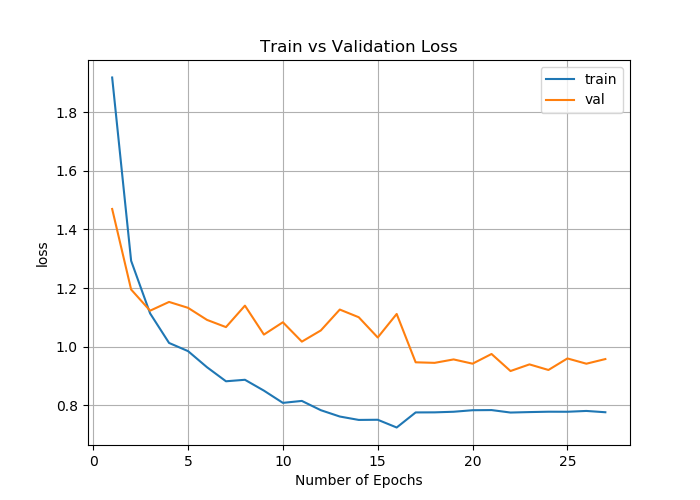}
        \caption{Loss based on L2}
    \end{subfigure}
    \caption{Effects of different regularization methods}
\end{figure}

\subsubsection{Optimizers}

We tested the effects of three different optimizers on our proposed model, RMSprop, Adam \cite{c25}, and SGD with momentum \cite{c26}. In figure 4, we can observe that the adaptive learning algorithm Adam reached convergence faster and on fewer epochs in comparison to SGD and RMSprop. In addition, we discovered that by using SGD and RMSprop optimizers, the model presents less overfit, but both optimizers were not able to reach higher accuracy than Adam. Thus, we decided to use Adam optimizer for further testing based on its performance. In table II, we present the parameters used for each of these three optimizers during training, as well as, their respective validation loss and accuracy.

In addition, a parameter that we adjusted often during training was the learning rate. We started out with a learning rate of 0.01, and decrease it by factors of 10 if we detected that our validation loss was beginning to overfit. From our experiments, we found out that the optimal learning rate for our purposes was 0.00001.

\begin{figure*}[ht!]
    \centering
    \begin{subfigure}[h]{0.3\textwidth}
        \centering
        \includegraphics[height=1.2in]{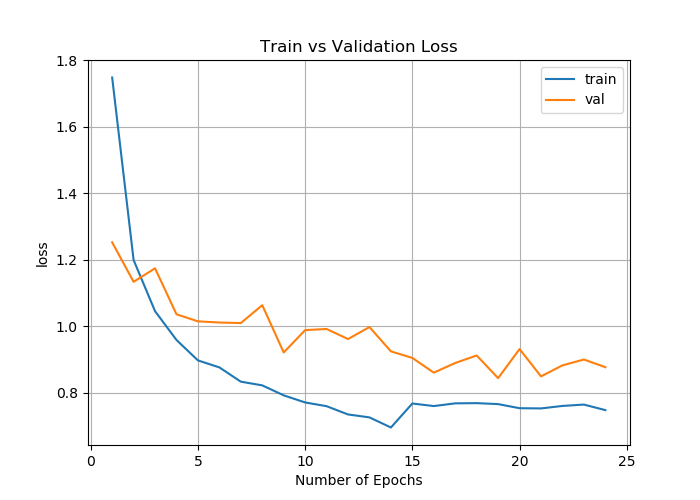}
        \caption{Adam Loss}
    \end{subfigure}%
    ~ 
    \begin{subfigure}[h]{0.3\textwidth}
        \centering
        \includegraphics[height=1.2in]{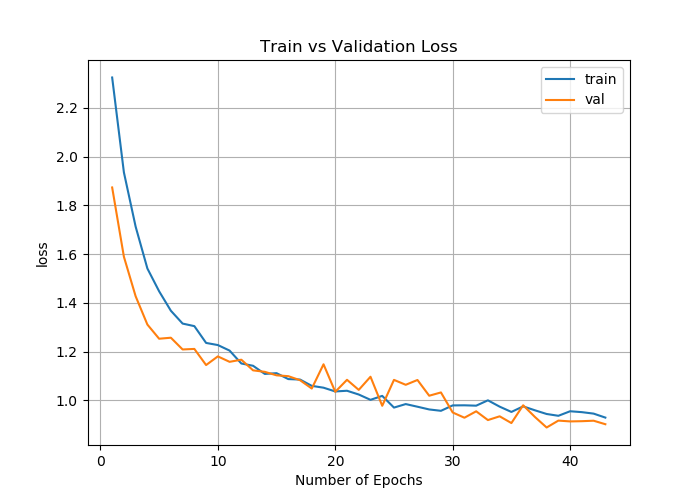}
        \caption{SGD Loss}
    \end{subfigure}%
    ~
    \begin{subfigure}[h]{0.3\textwidth}
        \centering
        \includegraphics[height=1.2in]{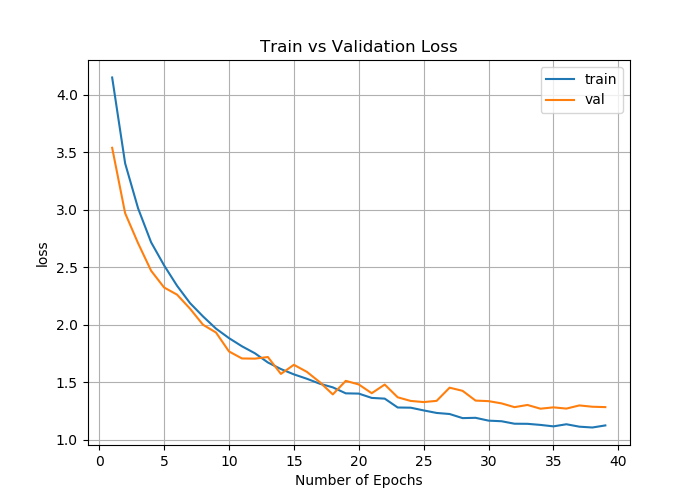}
        \caption{RMSprop Loss}
    \end{subfigure}
    
    \begin{subfigure}[h]{0.3\textwidth}
        \centering
        \includegraphics[height=1.2in]{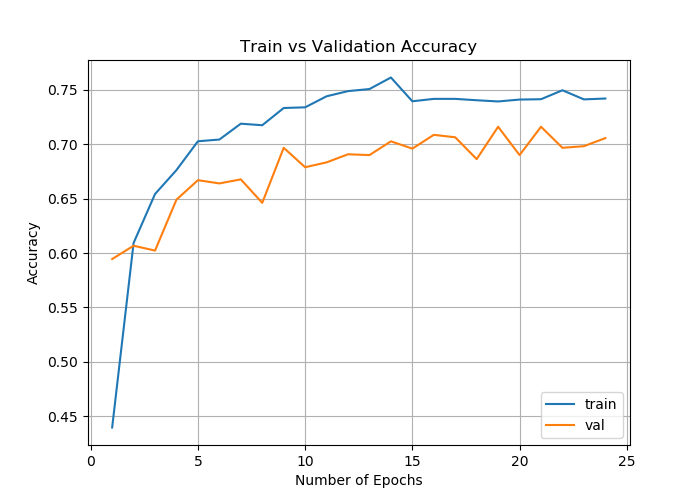}
        \caption{Adam Accuracy}
    \end{subfigure}%
    ~ 
    \begin{subfigure}[h]{0.3\textwidth}
        \centering
        \includegraphics[height=1.2in]{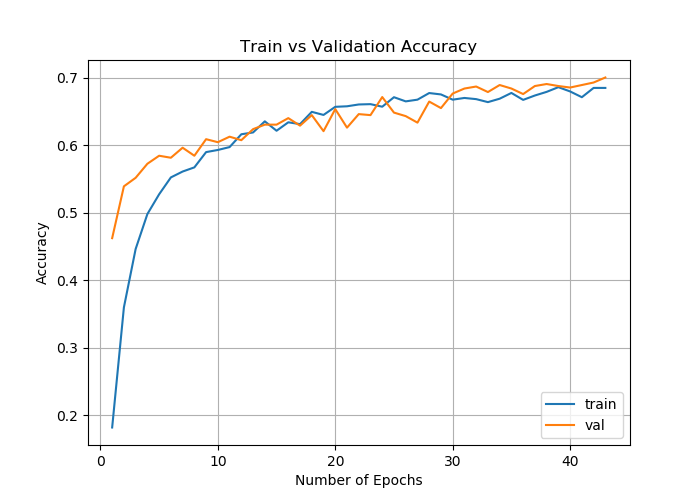}
        \caption{SGD Accuracy}
    \end{subfigure}%
    ~
    \begin{subfigure}[h]{0.3\textwidth}
        \centering
        \includegraphics[height=1.2in]{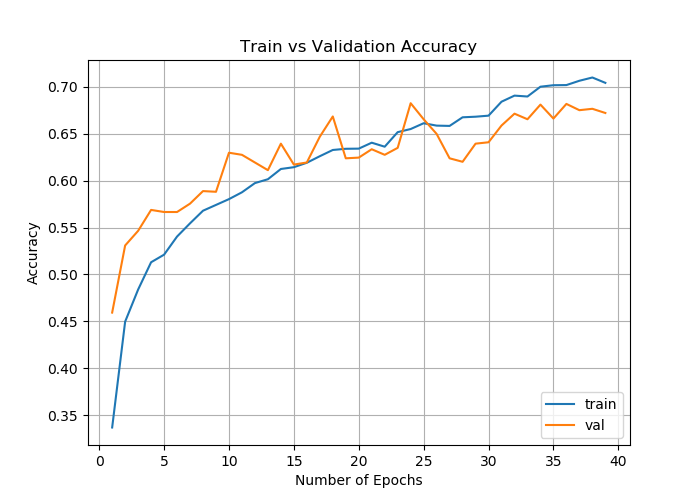}
        \caption{RMSprop Accuracy}
    \end{subfigure}
    \caption{Validation loss and accuracy for different optimizers}
\end{figure*}

\begin{table*}
\caption{Parameters used for different optimizers}
\label{table_example}
\begin{center}
\begin{tabular}{|c|c|c|}
\hline
Optimizers and parameters & Validation Loss & Validation Acc.\\
\hline
Adam($lr=0.00001, beta_1=0.9, beta_2=0.999, epsilon=1e-8$) & 0.8192  &0.7286 \\
\hline
SGD($lr=0.0001, momentum=0.9, nesterov=True$) & 0.9319 & 0.6907\\
\hline
RMSprop($lr=0.00001, rho=0.9$) & 1.2708 & 0.6810\\
\hline
\end{tabular}
\end{center}
\end{table*}

\subsubsection{Fine-tuning Layers}

The optimizer chosen for our second stage training of fine tuning was SGD, we found that this optimizer allows for slower convergence since it is not an adaptive algorithm such as Adam or RMSprop. Thus, SGD is perfect for fine tuning stage because it produces small updates to the already well-trained weights. Using an adaptive learning optimization algorithm may modify weights significantly, which means that some important features of both object and state classification could be loss. As a result, a significantly small learning rate must be used to prevent the possibility of overshooting an optimal minimum point of convergence. 

In our experiments, we decided to test the effects on freezing multiple layers. From the results in table III, we can observe that freezing up to the top two trainable blocks (freezing layers up to the 249th layer) gave us the best performance in terms of accuracy.

As it was shown in \cite{c2}, freezing more layers in our model started to decrease our accuracy, and also increased our training time significantly. The Inception V3 model was trained with the purpose of object detection. Hence, classifying object states bring a whole new set of challenges, and thus if we decided to train the network from scratch, it is possible that our network may be able to generalize better about the problem. However, due to time constraints, we decided to take advantage of transfer learning, which demonstrates that we can still obtain decent results that could be used as a starting point for finding a potential solution.

\begin{table}[ht]
\caption{Effects of freezing different layers}
\label{table2}
\begin{center}
\begin{tabular}{|c|c|c|c|}    
\hline
Freezing Layer & Trainable Parameters & Val Loss & Val Acc.\\
\hline
0 - 197 & 35,409,419 & 0.8486 & 0.7078\\
\hline
0 - 229 & 33,269,387 & 0.8348 & 0.7145 \\
\hline
0 - 249 & 31,572,363 & 0.8192  & 0.7286\\
\hline
\end{tabular}
\end{center}
\end{table}

\subsection{Results}

The best performing model was chosen to be tested on unseen data. Our model was able to obtain an accuracy of 72\% on validation data after training, and an accuracy of 69.4\% on unseen testing data. The plots for both loss and accuracy metrics during training are shown in figure 4(a) and 4(d).

\section{Discussion and Future work}

Although the  69.4\% accuracy reached by our model on unseen test data was decent, it was not as high as we expected to be. After a careful analysis on our cooking dataset, we found some misclassified images, as well as, ambiguous and multi-state images that could have contributed to a decrease on the overall performance of our model, and ultimately affect our object state classification. In figure 5, we presented a set of images that depicts the aforementioned problems in the dataset.

In \cite{c1}, it is mentioned that another factor that could have affected our classification results were the objects labeled as being in the "other" state. The "other" folder is composed of several cooking images including prepared dishes, raw meat, sandwiches, among others, that are not possible to represent with any of the other 10 states. In addition, this folder also contains a variety of images of ingredients in a combination of multiple states, which could lead to misclassifications. This suggests that performing joint detection as shown in \cite{c9} would allow our model to append all transformations of the cooking ingredient within the image, and could improve the overall performance of our model.

In conclusion, this paper proposed a modified fine-tuned Inception V3 model to recognize 11 different states of cooking ingredients (e.g. diced, peeled, floured, etc.), with the purpose of providing valuable insight and promising results into finding a potential solution for the object state recognition challenge, as well as, contributing in the advancement of robotic applications capable of performing daily-lives tasks such as cooking. Our model was able to achieve 72\% accuracy on validation data and 69.4\% accuracy on unseen test data. In a future study, a larger and more robust dataset will be used for better analysis and classification, as well as, an enhanced CNN architecture will be implemented to improve the performance on cooking-ingredient state detection. In the future, our purpose is to combine the different strategies on cooking-related problems such as cooking-recipe retrieval, object grasping and manipulation, and object state recognition to design robots capable of properly performing cooking tasks. 

\begin{figure*}[ht]
    \centering
    \begin{subfigure}[h]{0.2\textwidth}
        \includegraphics[height=1in]{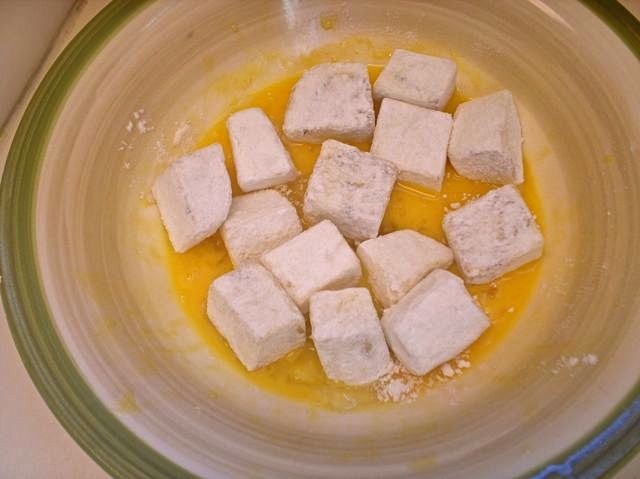}
        \caption{}
    \end{subfigure}%
    ~ 
    \begin{subfigure}[h]{0.2\textwidth}
        \includegraphics[height=1in]{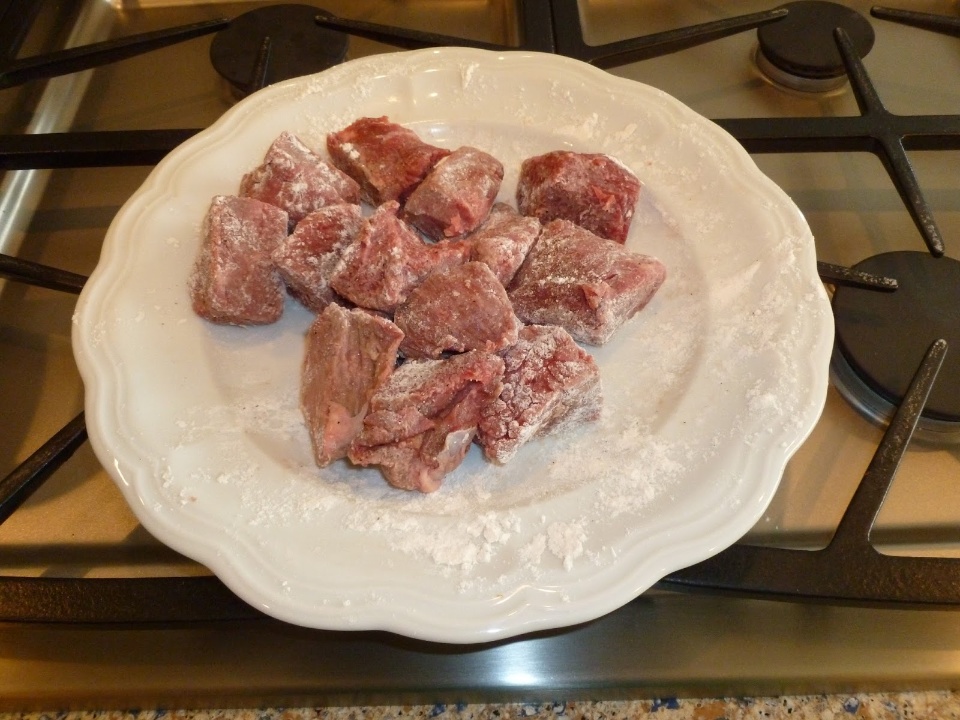}
        \caption{}
    \end{subfigure}%
    ~
    \begin{subfigure}[h]{0.2\textwidth}
        \includegraphics[height=1in, width=1.3in]{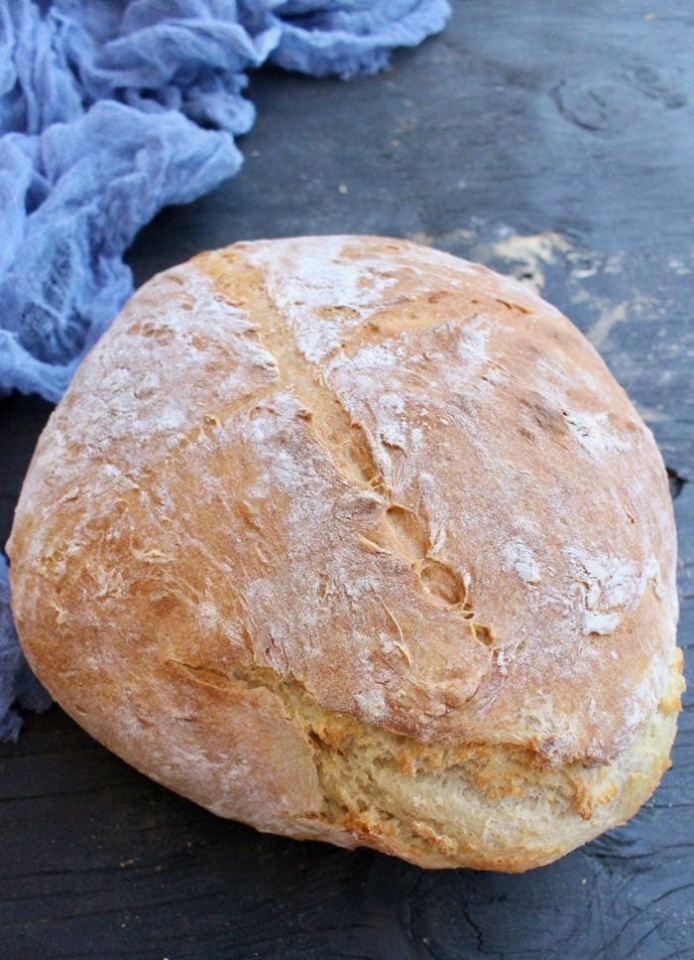}
        \caption{}
    \end{subfigure}%
    ~ 
    \begin{subfigure}[h]{0.2\textwidth}
        \includegraphics[height=1in]{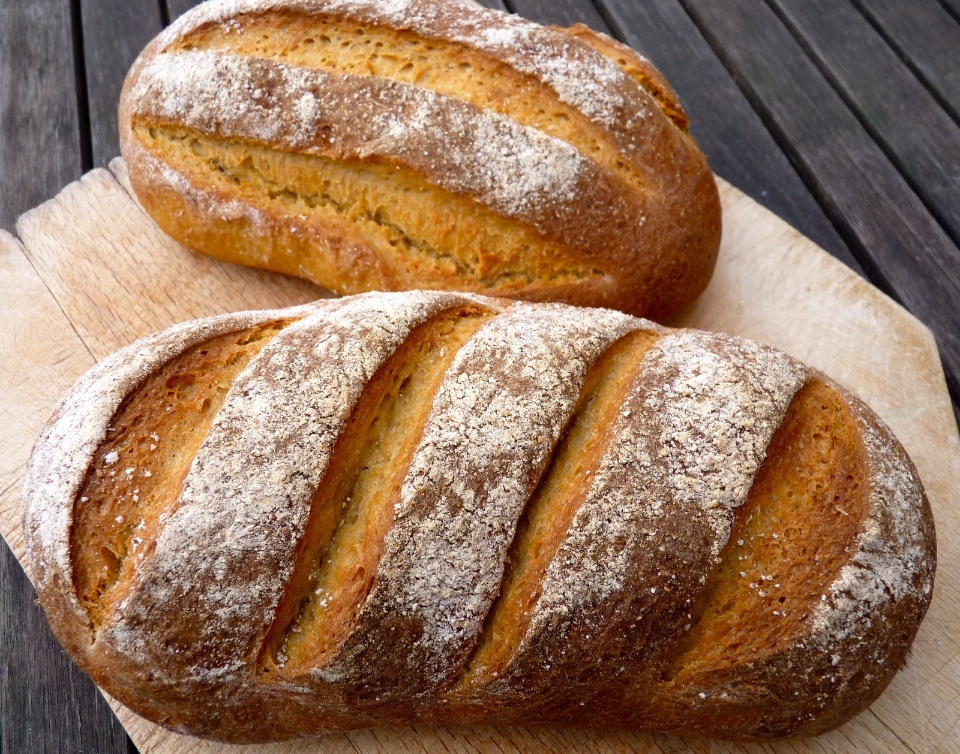}
        \caption{}
    \end{subfigure}
    \linebreak

    \begin{subfigure}[h]{0.2\textwidth}
        \includegraphics[height=1in, width=1.3in]{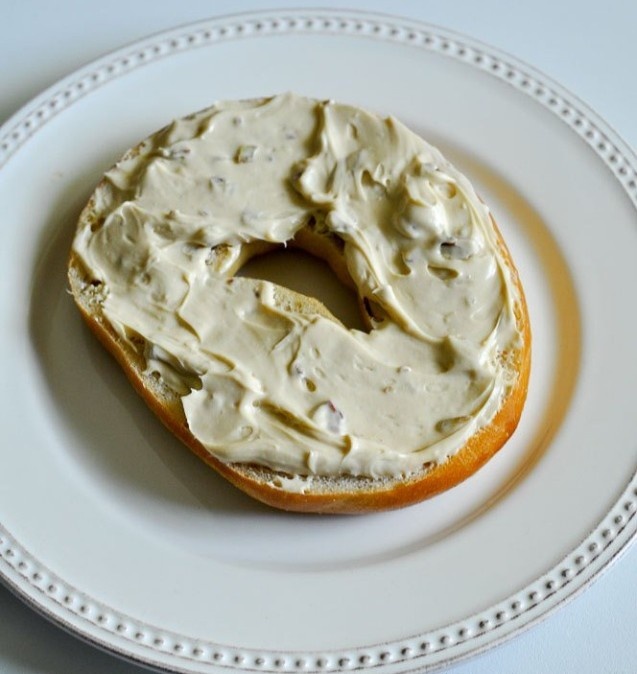}
        \caption{}
    \end{subfigure}%
    ~ 
    \begin{subfigure}[h]{0.2\textwidth}
        \includegraphics[height=1in]{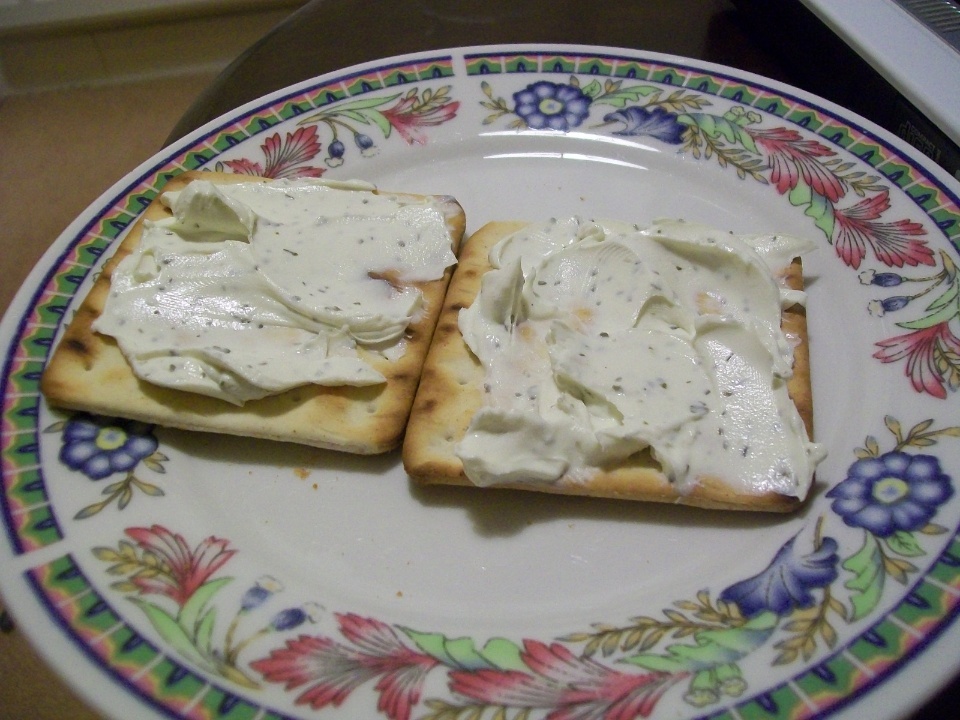}
        \caption{}
    \end{subfigure}%
    ~
    \begin{subfigure}[h]{0.2\textwidth}
        \includegraphics[height=1in, width=1.3in]{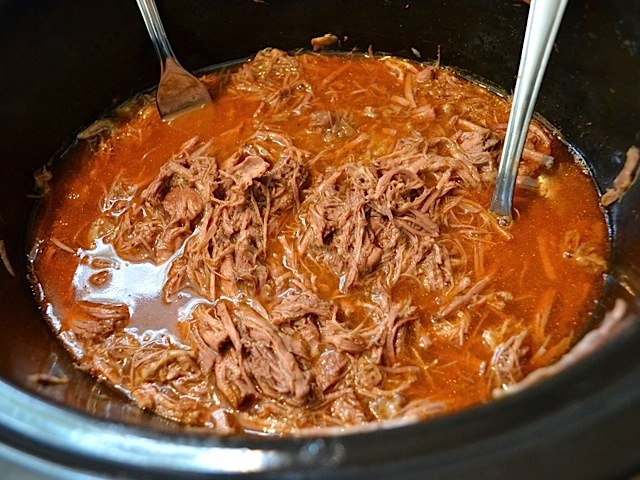}
        \caption{}
    \end{subfigure}%
    ~ 
    \begin{subfigure}[h]{0.2\textwidth}
        \includegraphics[height=1in, width=1.3in]{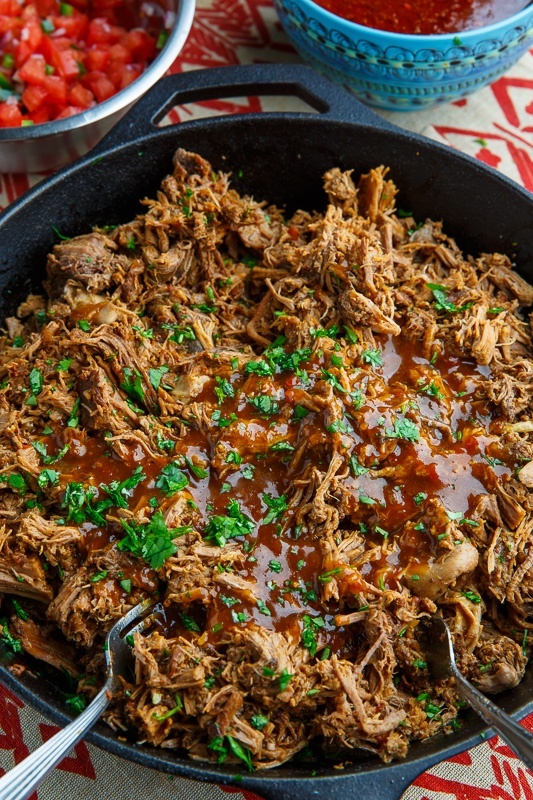}
        \caption{}
    \end{subfigure}
    \linebreak
    
    \begin{subfigure}[h]{0.2\textwidth}
        \includegraphics[height=1in, width=1.3in]{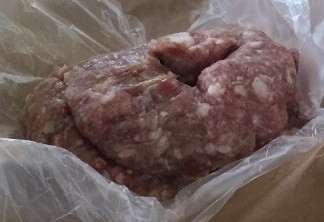}
        \caption{}
    \end{subfigure}
    
    \caption{Example of multi-states: (a-b) diced and floured, (c-d) whole and floured. Example of similar images but different labels: (e) labeled as creamy and (f) labeled as other, (g) labeled as julienne and (h) labeled as other. Example of misclassified image: (i) original label should be other, classified as creamy. }
\end{figure*}

\bibliographystyle{ieeetr}

\addtolength{\textheight}{-12cm}   





\end{document}